\documentclass[11pt,a4paper]{article}
\usepackage[hyperref]{naaclhlt2018}
\usepackage{times}
\usepackage{latexsym}
\usepackage{amsmath}
\usepackage{multirow}
\usepackage{graphicx}
\usepackage{xspace}
\usepackage[draft,textsize=footnotesize]{todonotes}
\usepackage{enumitem}

\usepackage[draft,textsize=footnotesize]{todonotes}
 
\newcommand{\acropt}{{\sc PT16}\xspace}
\newcommand{\acroxu}{{\sc Xu12}\xspace}
\newcommand{\acroh}{{\sc H14}\xspace}
\newcommand{\acrohe}{{\sc He15}\xspace}
\newcommand{\acrocorp}{{\sc GYAFC}\xspace}
\usepackage{url}

\aclfinalcopy 

\title{Dear Sir or Madam, May I Introduce the GYAFC Dataset: \\ Corpus, Benchmarks and Metrics for Formality Style Transfer}

\author{Sudha Rao \\
  University of Maryland, College Park\thanks{This research was performed when the first author was at Grammarly.} \\
  {\tt raosudha@cs.umd.edu} \\\And
  Joel Tetreault \\
  Grammarly \\
  {\tt joel.tetreault@grammarly.com} \\}

\date{}

\begin{document}

\maketitle

\begin{abstract}
Style transfer is the task of automatically transforming a piece of text in one particular style into another.  
A major barrier to progress in this field has been a lack of training and evaluation datasets, as well as benchmarks and automatic metrics.  In this work, we create the largest corpus for a particular stylistic transfer (formality) and show that techniques from the machine translation community can serve as strong baselines for future work.  We also discuss challenges of using automatic metrics.
\end{abstract}

\section{Introduction}

One key aspect of \emph{effective communication} is the accurate expression of the style or tone of some content. For example, writing a more \textit{persuasive email} in a marketing position could lead to increased sales; writing a more \textit{formal email} when applying for a job could lead to an offer; and writing a more \textit{polite note} to your future spouse's parents, may put you in a good light. Hovy \shortcite{hovy1987generating} argues that by varying the style of a text, people convey more information than is present in the literal meaning of the words. One particularly important dimension of style is formality \cite{heylighen1999formality}.  
Automatically changing the style of a given content to make it more formal can be a useful addition to any writing assistance tool.

In the field of style transfer, to date, the only available dataset has been for the transformation of modern English to Shakespeare, and it led to the application of phrase-based machine translation (PBMT) \cite{xu2012paraphrasing} and neural machine translation (NMT) \cite{jhamtani2017shakespearizing} models to the task. The lack of an equivalent or larger dataset for any other form of style transfer has blocked progress in this field. Moreover, prior work has mainly borrowed metrics from machine translation (MT) and paraphrase communities for evaluating style transfer. However, it is not clear if those metrics 
are the best ones to use for this task.  
In this work, we address these issues through the following three contributions:
\begin{itemize}[noitemsep,nolistsep]
\item \textbf{Corpus:} We present Grammarly's Yahoo Answers Formality Corpus (\acrocorp), the largest dataset for any style containing a total of 110K informal / formal sentence pairs.  
Table~\ref{tab:informal_formal_examples} shows sample sentence pairs. 
\item \textbf{Benchmarks:} We introduce a set of learning models for the task of formality style transfer. Inspired by work in low resource MT, we adapt existing PBMT and NMT approaches for our task and show that they can serve as strong benchmarks for future work.
\item \textbf{Metrics:} In addition to MT and paraphrase metrics, we evaluate our models along three axes: \textit{formality}, \textit{fluency} and \textit{meaning preservation} using existing automatic metrics. We compare these  metrics with their human judgments and show there is much room for further improvement. 
\end{itemize}
\begin{table}[h]
\small
\begin{tabular}{l}
\hline
Informal: \textit{I'd say it is punk though.} \\
Formal: \hspace{1mm} \textit{However, I do believe it to be punk.}\\
\hline
Informal: \textit{Gotta see both sides of the story.} \\
Formal: \hspace{1mm} \textit{You have to consider both sides of the story.}\\
\hline
\end{tabular}
\caption{Informal sentences with formal rewrites.}\label{tab:informal_formal_examples}
\vspace*{-2mm}
\end{table}
In this paper, we primarily focus on the \textit{informal} to \textit{formal} direction since we collect our dataset for this direction. However, we evaluate our models on the \textit{formal} to \textit{informal} direction as well.\footnote{Results are in the supplementary material.}
All data, model outputs, and evaluation results have been made public\footnote{\url{https://github.com/raosudha89/GYAFC-corpus}} in the hope that they will encourage more research into style transfer.


In the following two sections we discuss related work and the \acrocorp dataset.  In \S\ref{model}, we detail our rule-based and  MT-based approaches. In \S\ref{evaluation}, we describe our human and automatic metric based evaluation.  In \S\ref{results}, we describe the results of our models using both human and automatic evaluation and discuss how well the automatic metrics correlate with human judgments. 

%
%
%

\section{Related Work}

\textbf{Style Transfer with Parallel Data:}
Sheikha and Inkpen \shortcite{sheikha2011generation} collect pairs of formal and informal words and phrases from different sources and use a natural language generation system to generate informal and formal texts by replacing lexical items based on user preferences. Xu et al. \shortcite{xu2012paraphrasing} (henceforth \acroxu) was one of the first works to treat style transfer as a sequence to sequence task. They generate a parallel corpus of 30K sentence pairs by scraping the modern translations of Shakespeare plays and train a PBMT system to translate from modern English to Shakespearean English.\footnote{https://github.com/cocoxu/Shakespeare} More recently, Jhamtani et al. \shortcite{jhamtani2017shakespearizing} show that a copy-mechanism enriched sequence-to-sequence neural model outperforms \acroxu on the same set. In text simplification, the availability of parallel data extracted from English Wikipedia and Simple Wikipedia \cite{zhu2010monolingual} led to the application of PBMT \cite{wubben2012sentence} and more recently NMT \cite{wang2016text} models. 
We take inspiration from both the PBMT and NMT models and apply several modifications to these approaches for our task of transforming the formality style of the text. \\\\
\textbf{Style Transfer without Parallel Data:} Another direction of research directly controls certain attributes of the generated text {\em without} using parallel data.   Hu et al. \shortcite{hu2017toward} control the sentiment and the tense of the generated text by learning a disentangled latent representation in a neural generative model. 
Ficler and Goldberg \shortcite{ficler2017controlling} control several linguistic style aspects simultaneously by conditioning a recurrent neural network language model on specific style (professional, personal, length) and content (theme, sentiment) parameters.  
Under NMT models, Sennrich et al. \shortcite{sennrich2016controlling} control the politeness of the translated text via side constraints, 
Niu et al. \shortcite{niu2017study} control the level of formality of MT output by selecting phrases of a requisite formality level from the k-best list during decoding.  
In the field of text simplification, more recently, \newcite{xu2016optimizing} learn large-scale paraphrase rules using bilingual texts whereas \newcite{kajiwara2016building} build a monolingual parallel corpus using sentence similarity based on alignment between word embeddings.  
Our work differs from these methods in that we mainly address the question of how much leverage we can derive by collecting a large amount of informal-formal sentence pairs and build models that learn to transfer style directly using this parallel corpus. \\\\
\textbf{Identifying Formality:} There has been previous work on detecting formality of a given text at the lexical level \cite{brooke2010automatic,lahiri2011informality,brooke2014supervised,pavlick2015inducing}, at the sentence level \cite{pavlick2016empirical} and at the document level \cite{sheikha2010automatic,peterson2011email,mosquera2012smile}. In our work, we reproduce the sentence-level formality classifier introduced in Pavlick and Tetreault \shortcite{pavlick2016empirical} (\acropt) to  extract informal sentences for \acrocorp creation and to automatically evaluate system outputs.
\\\\
\textbf{Evaluating Style Transfer:} The problem of style transfer falls under the category of natural language generation tasks such as machine translation, paraphrasing,  etc. 
Previous work on style transfer \cite{xu2012paraphrasing,jhamtani2017shakespearizing,niu2017study,sennrich2016controlling} has re-purposed the MT metric BLEU \cite{papineni2002bleu} and the paraphrase metric PINC \cite{chen2011collecting} for evaluation. Additionally, \acroxu introduce three new automatic style metrics based on cosine similarity, language model and logistic regression that measure the degree to which the output matches the target style. Under human based evaluation, on the other hand, there has been work on a more fine grained evaluation where human judgments were separately collected for adequacy, fluency and style \cite{xu2012paraphrasing,niu2017study}. In our work, we conduct a more thorough evaluation where we evaluate model outputs on the three criteria of \textit{formality}, \textit{fluency} and \textit{meaning} using both automatic metrics and human judgments.

%
%
%

\section{\acrocorp Dataset}\label{dataset}

\subsection{Creation Process}
Yahoo Answers,\footnote{\url{https://answers.yahoo.com/answer}} a question answering forum, contains a large number of informal sentences and allows redistribution of data. Hence, we use the Yahoo Answers L6 corpus\footnote{\url{https://webscope.sandbox.yahoo.com/catalog.php?datatype=l}} to create our \acrocorp dataset of informal and formal sentence pairs. In order to ensure a uniform distribution of data, we remove sentences that are questions, contain URLs, and are shorter than 5 words or longer than 25. After these preprocessing steps, 40 million sentences remain. The Yahoo Answers corpus consists of several different domains like \textit{Business, Entertainment \& Music, Travel, Food,\textit} etc. 
\acropt show that the formality level varies significantly across different genres. 
In order to control for this variation, we work with two specific domains that contain the most informal sentences and show results on training and testing within those categories.  
We use the formality classifier from \acropt to identify informal sentences. We train this classifier on the \textit{Answers} genre of the \acropt corpus which consists of nearly 5,000 randomly selected sentences from Yahoo Answers manually annotated on a scale of -3 (very informal) to 3 (very formal).\footnote{\url{http://www.seas.upenn.edu/~nlp/resources/formality-corpus.tgz}} 
We find that the domains of \textit{Entertainment \& Music} and \textit{Family \& Relationships} contain the most informal sentences and create our \acrocorp dataset using these domains.  Table~\ref{corpus_statistics} shows the number of formal and informal sentences in all of Yahoo Answers corpus and within the two selected domains. Sentences with a score less than 0 are considered as informal and sentences with a score greater than 0 are considered as formal.

\begin{table}
\small
\begin{tabular}{l c  c  c}
 Domain & Total & Informal & Formal \\
\hline
All Yahoo Answers & 40M & 24M & 16M \\
Entertainment \& Music & 3.8M & 2.7M & 700K \\
Family \& Relationships & 7.8M & 5.6M & 1.8M \\
\end{tabular}
\caption{Yahoo Answers corpus statistics}\label{corpus_statistics}
\end{table}

Next, we randomly sample a subset of 53,000 informal sentences each from the \textit{Entertainment \& Music} (E\&M) and \textit{Family \& Relationships} (F\&R) categories and collect one formal rewrite per sentence using Amazon Mechanical Turk.  The workers are presented with detailed instructions, as well as examples. 
To ensure quality control, four experts, two of which are the authors of this paper, reviewed the rewrites of the workers and rejected those that they felt did not meet the required standards. They also provided the workers with reasons for rejection so that they would not repeat the same mistakes. Any worker who repeatedly performed poorly was eventually blocked from doing the task. 
We use this train set to train our models for the style transfer tasks in both directions.

\begin{table}
\footnotesize
\begin{tabular}{l  c | c  c | c  c}
 & & \multicolumn{2}{l|}{\textit{Informal to Formal}} & \multicolumn{2}{l}{\textit{Formal to Informal}}\\
 \hline
 & Train & Tune & Test & Tune & Test\\
 \hline
E\&M & 52,595 & 2,877 & 1,416 & 2,356 & 1,082\\
F\&R & 51,967 & 2,788 & 1,332 & 2,247 & 1,019\\
\end{tabular}
\caption{\acrocorp dataset statistics}\label{dataset_statistics}
\end{table}

Since we want our tune and test sets to be of higher quality compared to the train set, we recruit a set of 85 expert workers for this annotation who had a 100\% acceptance rate for our task and who had previously done more than 100 rewrites. Further, we collect multiple references for the tune/test set to adapt PBMT tuning and evaluation techniques to our task. We collect four different rewrites per sentence using our expert workers by randomly assigning sentences to the experts until four rewrites for each sentence are obtained.\footnote{Thus, note that the four rewrites are not from the same four workers for each sentence} To create our tune and test sets for the \textit{informal} to \textit{formal} direction, we sample an additional 3,000 informal sentences for our tune set and 1,500 sentences for our test set from each of the two domains.  

To create our tune and test sets for the \textit{formal} to \textit{informal} direction, we start with the same tune and test split as the first direction. For each formal rewrite\footnote{Out of four, we pick the one with the most edit distance with the original informal. Rationale explained in Section~\ref{data_analysis}} from the first direction, we collect three different informal rewrites using our expert workers as before. These three informal rewrites along with the original informal sentence become our set of four references for this direction of the task. 
Table~\ref{dataset_statistics} shows the exact number of sentences in our train, tune and test sets.

%
%
%

\subsection{Analysis}\label{data_analysis}
The following quantitative and qualitative analyses are aimed at characterizing the changes between the original informal sentence and its formal rewrite in the \acrocorp train split.\footnote{We observe similar patterns on the tune and test set.}  
We present our analysis here on only the E\&M domain data since we observe similar patterns in F\&R.




\begin{figure}[h]
	\begin{minipage}[b]{0.5\textwidth}
	\includegraphics[scale=0.45]{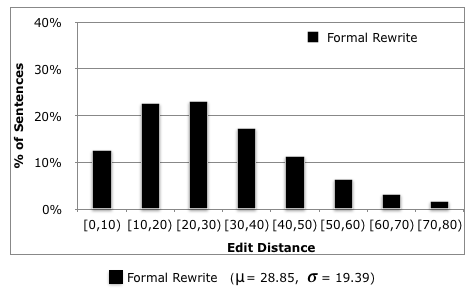}
    \caption{Percentage of sentences binned according to formality score in train set of E\&M}\label{fig:EM_EditDist_Train}
	\end{minipage}
    \begin{minipage}[b]{0.5\textwidth}
    \includegraphics[scale=0.45]{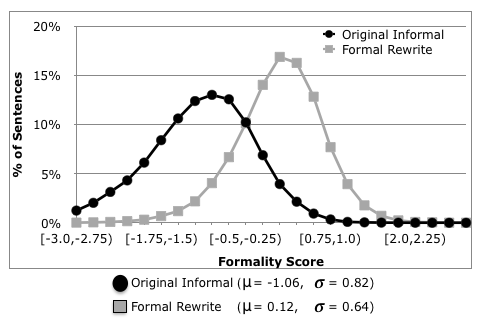}
    \caption{Percentage of sentences binned according to formality score in train set of E\&M}\label{fig:EM_Formality_Train}
    \end{minipage}
\end{figure}

\noindent {\bf Quantitative Analysis:} While rewriting sentences more formally, humans tend to make a wide range of lexical/character-level edits. In Figure~\ref{fig:EM_EditDist_Train}, we plot the distribution of the character-level Levenshtein edit distance between the original informal and the formal rewrites in the train set and observe a standard deviation of $\sigma = 19.39$ with a mean $\mu = 28.85$.  
Next, we look at the difference in the formality level of the original informal and the formal rewrites in \acrocorp. We find that the classifier trained on the \textit{Answers} genre of \acropt dataset correlates poorly (Spearman $\rho$ = 0.38) with human judgments when tested on our domain specific 
datasets.   
Hence, we collect formality judgments on a scale of -3 to +1, similar to \acropt, for an additional 5000 sentences each from both domains and obtain a formality classifier with higher correlation (Spearman $\rho$ = 0.56). 
We use this retrained classifier for our evaluation in \S\ref{evaluation} as well. 

In Figure~\ref{fig:EM_Formality_Train}, we plot the distribution of the formality scores on the original informal sentence and their formal rewrites in the train set and observe an increase in the mean formality score as we go from informal ($-1.06$) to formal rewrites ($0.12$).
As compared to edit distance and formality, we observe a much lower variation in sentence lengths 
with the mean slightly increasing from informal ($11.93$) to their formal rewrites ($12.56$) in the train set.
\\\\
\noindent {\bf Qualitative Analysis:} To understand what stylistic choices differentiate formal from informal text, we perform an analysis similar to \acropt and look at 50 rewrites from both domains and record the frequency of the types of edits that workers made when creating a more formal sentence.
\footnote{Examples of edits in supplementary material.}
In contrast to \acropt, we observe a higher percentage of phrasal paraphrases (47\%), edits to punctuations (40\%) and expansion of contractions (12\%). This is reflective of our sentences coming from very informal domains of Yahoo Answers. Similar to \acropt, we also observe capitalization (46\%) and normalization (10\%).
%
%
%
\section{Models}\label{model}

We experiment with three main classes of approaches: a rule-based approach, PBMT and NMT. Inspired by work in low resource machine translation, we apply several modifications to the standard PBMT and NMT models and create a set of strong benchmarks for the style transfer community.
We apply these models to both directions of style transfer: \textit{informal} to \textit{formal} and \textit{formal} to \textit{informal}. In our description, we refer to the two styles as \textit{source} and \textit{target}. 
We summarize the models below and direct the reader to supplementary material for further detail. 

\subsection{Rule-based Approach}\label{rule_based}
Corresponding to the category of edits described in \S\ref{data_analysis}, 
we develop a set of rules to automatically make an informal sentence more formal where we capitalize first word and proper nouns, remove repeated punctuations, handcraft a list of expansion for contractions etc. For the \textit{formal}  to \textit{informal} direction, we design a similar set of rules in the opposite direction.

%
%
\subsection{Phrase-based Machine Translation}\label{smt_models}
Phrased-based machine translation 
models have had success in the fields of machine translation, style transfer (\acroxu) and text simplification \cite{wubben-vandenbosch-krahmer:2012:ACL2012,xu2016optimizing}. 
Inspired by work in low resource machine translation, we use a combination of training regimes to develop our model.  We train on the output of the rule-based approach when applied to \acrocorp.  This is meant to force the PBMT model to learn generalizations {\em outside} the rules.  To increase the data size, we use self-training \cite{ueffing2006self}, where we use the PBMT model to translate the large number of in-domain sentences from \acrocorp belonging to the the source style and use the resultant output to retrain the PBMT model.  Using sub-selection, we only select rewrites that have an Levenshtein edit distance of over 10 characters when compared to the source to encourage the model to be less conservative.  Finally, we upweight the rule-based \acrocorp data via duplication \cite{sennrich-haddow-birch:2016:WMT}.  For our experiments, we use Moses \cite{koehn2007moses}. We train a 5-gram language model using KenLM \cite{Heafield-estimate}, and use target style sentences from \acrocorp and the sub-sampled target style sentences from out-of-domain Yahoo Answers, as in Moore and Lewis \shortcite{moore2010intelligent}, to create a large language model.

\subsection{Neural Machine Translation}\label{nmt_models}
While encoder-decoder based neural network models have become quite successful for MT\cite{sutskever2014sequence,bahdanau2014neural,cho2014properties},  
the field of style transfer, has not yet been able to fully take advantage of these advances owing to the lack of availability of large parallel data.  With \acrocorp we can now show how well NMT techniques fare for style transfer. We experiment with three NMT models:
\\\\
\noindent {\bf NMT baseline:} Our baseline model is a bi-directional LSTM \cite{hochreiter1997long} encoder-decoder model with attention \cite{bahdanau2014neural}.\footnote{Details are in the supplementary material.}
We pretrain the input word embeddings on Yahoo Answers using GloVE \cite{pennington2014glove}.
As in our PBMT based approach, we train our NMT baseline model on the output of the rule-based approach when applied to \acrocorp.
\\\\
\noindent \textbf{NMT Copy:} Jhamtani et al., \shortcite{jhamtani2017shakespearizing} introduce a copy-enriched NMT model for style transfer to better handle stretches of text which should not be changed. We incorporate this mechanism into our NMT Baseline.
\\\\
\noindent \textbf{NMT Combined:} The size of our parallel data is smaller than the size typically used to train NMT models. Motivated by this fact, we propose several variants to the baseline models that we find helps minimize this issue. We augment the data used to train  NMT Copy via two techniques:  1) we run the PBMT model on additional source data, and 2) we use back-translation \cite{P16-1009} of the PBMT model to translate the large number of in-domain target style sentences from \acrocorp.  To balance the over one million artificially generated pairs from the respective techniques, we upweight the rule-based \acrocorp data via duplication.\footnote{Training data sizes for different methods are summarized in the supplementary material.}

\section{Evaluation}\label{evaluation}

As discussed earlier, there has been very little research
into best practices for style transfer evaluation.  Only a few works have included a human evaluation \cite{xu2012paraphrasing,jhamtani2017shakespearizing}, and automatic evaluations have employed BLEU or PINC \cite{xu2012paraphrasing,chen2011collecting}, which have been borrowed from other fields and not vetted for this task.
In our work, we conduct a more thorough and detailed evaluation using both humans and automatic metrics to assess transformations.  Inspired by work in the paraphrase community \cite{callisonburch:2008:EMNLP}, we solicit ratings on how formal, how fluent and how meaning-preserving a rewrite is.
Additionally, we look at the correlation between the human judgments and the automatic metrics. 

\subsection{Human-based Evaluation}\label{human_judgments}
We perform human-based evaluation to assess model outputs on the four criteria: \textit{formality}, \textit{fluency}, \textit{meaning} and \textit{overall}. For a subset of 500 sentences from the test sets of both \textit{Entertainment \& Music} and \textit{Family \& Relationship} domains, we collect five human judgments per sentence per criteria using Amazon Mechanical Turk as follows:
\\\\
\noindent\textbf{Formality:} Following \acropt, workers rate the formality of the source style sentence, the target style reference rewrite and the target style model outputs on a discrete scale of -3 to +3 described as: 
\textit{-3: Very Informal, -2: Informal, -1: Somewhat Informal, 0: Neutral, 1: Somewhat Formal, 2: Formal and 3: Very Formal}.  
\normalsize
\\\\
\noindent\textbf{Fluency:} 
Following Heilman et al. \shortcite{heilman-EtAl:2014:P14-2}, workers rate the fluency of the source style sentence, the target style reference rewrite and the target style model outputs on a discrete scale of 1 to 5 described as: \textit{5: Perfect, 4: Comprehensible, 3: Somewhat Comprehensible, 2: Incomprehensible}. We additionally provide an option of \textit{1: Other} for sentences that are incomplete or just a fragment.
\\\\
\noindent\textbf{Meaning Preservation:} Following the annotation scheme developed for the Semantic Textual Similarity (STS) dataset \cite{agirre2016semeval}, given two sentences i.e. the source style sentence and the target style reference rewrite or the target style model output, workers rate the meaning similarity of the two sentences on a scale of 1 to 6 described as: \textit{6: Completely equivalent, 5: Mostly equivalent, 4: Roughly equivalent, 3: Not equivalent but share some details, 2: Not equivalent but on same topic, 1: Completely dissimilar}.
\\\\
\noindent\textbf{Overall Ranking:} In addition to the fine-grained human judgments, we collect judgments to assess the overall ranking of the systems. Given the original source style sentence, the target style reference rewrite and the target style model outputs, we ask workers to rank the rewrites in the order of their overall formality, taking into account both fluency and meaning preservation. We then rank the model using the equation below:
\begin{equation}
rank(model) = \frac{1}{|S|}\sum_{s \in S} \frac{1}{|J|}\sum_{j \in J} rank(s_{model}, j)
\end{equation}
where, $model$ is the one of our models, $S$ is a subset of 500 test set sentences, $J$ is the set of five judgments, $s_{model}$ is the model rewrite for sentence $s$, and $rank(s_{model}, j)$ is the rank of $s_{model}$ in judgment $j$.

The two authors of the paper reviewed these human judgments and found that in majority of the cases the annotations looked correct. But as is common in any such crowdsourced data collection process, there were some errors, especially in the overall ranking of the systems.

\subsection{Automatic Metrics}\label{automatic_metrics}
We cover each of the human evaluations with a corresponding automatic metric:\\\\
\textbf{Formality:} We use the formality classifier described in \acropt. 
We find that the classifier trained on the \textit{answers} genre of \acropt dataset does not perform well when tested on our datasets. Hence, we collect formality judgments for an additional 5000 sentences and use the formality classifier re-trained on this in-domain data. \\\\
\textbf{Fluency:} We use the reimplementation\footnote{https://github.com/cnap/grammaticality-metrics/tree/master/heilman-et-al} of Heilman et al. \shortcite{heilman-EtAl:2014:P14-2} (\acroh in Table~\ref{tab:results_EM}) which is a statistical model for predicting the grammaticality of a sentence on a scale of 0 to 4 previously shown to be effective for other generation tasks like grammatical error correction \cite{napoles-sakaguchi-tetreault:2016:EMNLP2016}. \\\\
\textbf{Meaning Preservation:} Modeling semantic similarity at a sentence level is a fundamental language processing task, and one that is a wide open field of research. 
Recently, He et al., \shortcite{he2015multi} (\acrohe in Table~\ref{tab:results_EM}) developed a convolutional neural network based sentence similarity measure. 
We use their off-the-shelf implementation\footnote{https://github.com/castorini/MP-CNN-Torch} to train a model on the STS and use it to measure the meaning similarity between the original source style sentence and its target style rewrite (both reference and model outputs). \\\\
\textbf{Overall Ranking:} We experiment with BLEU \cite{papineni2002bleu} and  PINC \cite{chen2011collecting} as both were used in prior style evaluations, as well as TERp \cite{snover2009fluency}.

\begin{table*}[htbp]
\center
\small
\begin{tabular}{l |c c| c c| c c| c c| c c c}
& \multicolumn{2}{c|}{Formality} & \multicolumn{2}{c|}{Fluency} & \multicolumn{2}{c}{Meaning} & \multicolumn{2}{|c} {Combined} & \multicolumn{3}{|c}{Overall}\\
Model &  Human & PT16 & Human & \acroh & Human & \acrohe & Human & Auto & BLEU & TERp & PINC\\
\hline
\hline
\textit{Original Informal} &\textit{ -1.23} & \textit{-1.00} & \textit{3.90} & \textit{2.89}  & -- & -- & -- & -- & \textit{50.69} & \textit{0.35} & \textit{0.00} \\
Formal Reference & 0.38 & 0.17 & 4.45 & 3.32 & 4.57 & 3.64 & 5.68 & 4.67 & 100.0 & 0.37 & 69.79\\
\hline
\hline
Rule-based  & -0.59 & -0.34 & 4.00 & 3.09 & \textbf{4.85} & \textbf{4.41} & 5.24 & 4.69 & 61.38 & 0.27 & 26.05\\
PBMT & -0.19* & 0.00* & 3.96 & 3.28* & 4.64* & 4.19* & 5.27 & 4.82* & 67.26* & \textbf{0.26} & 44.94*\\
NMT Baseline & \textbf{0.05}* & 0.07* & 4.05 & \textbf{3.52}* & 3.55* & 3.89* & 4.96* & \textbf{4.84}* & 56.61 & 0.38* & \textbf{56.92}*\\
NMT Copy & 0.02* & \textbf{0.10}* & 4.07 & 3.45* & 3.48* & 3.87* & 4.93* & 4.81* & 58.01 & 0.38* & 56.39*\\
NMT Combined & -0.16* & 0.00* & \textbf{4.09}* & 3.27* & 4.46* & 4.20* & \textbf{5.32}* & 4.82* & \textbf{67.67}* & \textbf{0.26} & 43.54*\\
\hline
\end{tabular}
\caption{Results of models on 500 test sentences from E\&M for \textit{informal} to \textit{formal} task evaluated using human judgments and automatic metrics for three criteria of evaluation: formality, fluency and meaning preservation. Scores marked with * are significantly different from the rule-based scores with $p < 0.001$.}\label{tab:results_EM}
\end{table*}


\begin{figure}[h]
	\begin{minipage}[b]{0.5\textwidth}
	\includegraphics[scale=0.33]{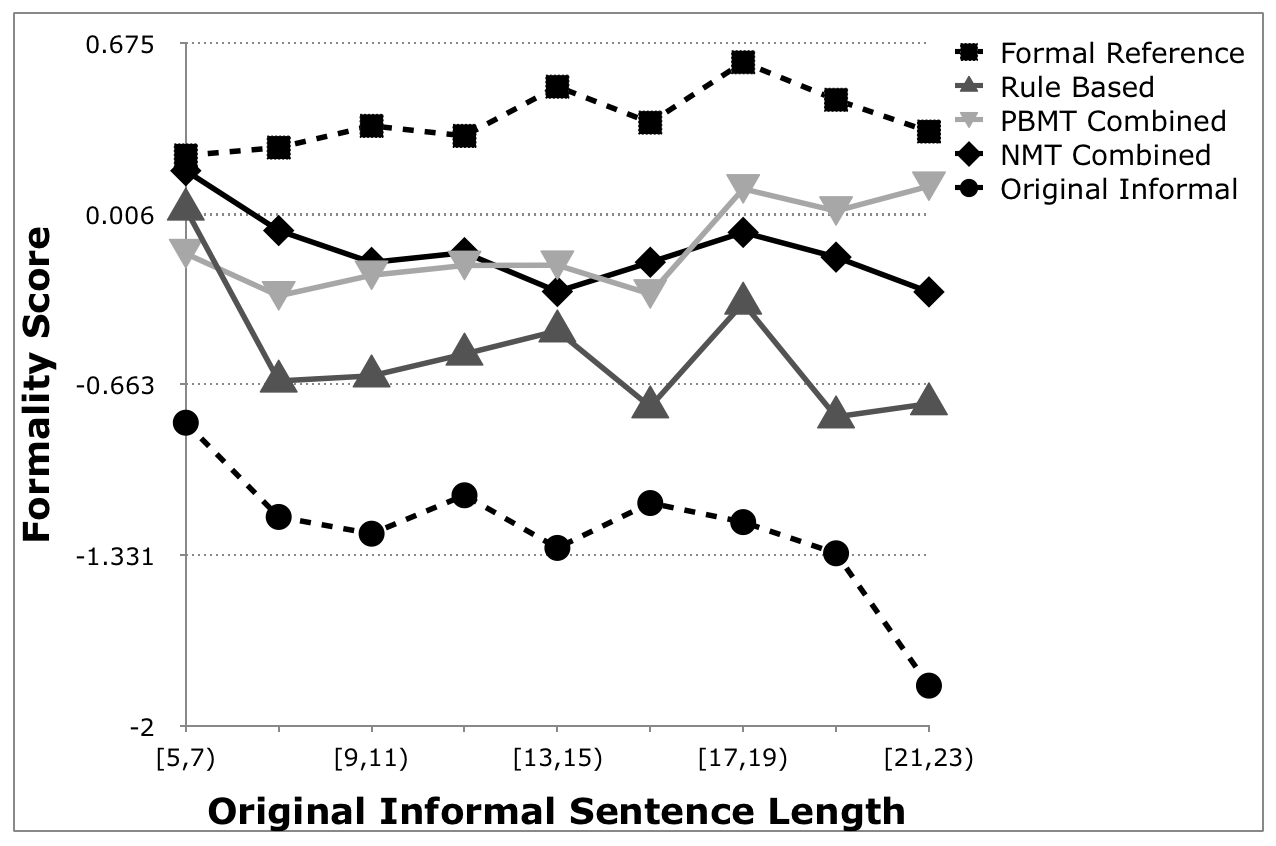}
	\end{minipage}
    \begin{minipage}[b]{0.5\textwidth}
    \includegraphics[scale=0.33]{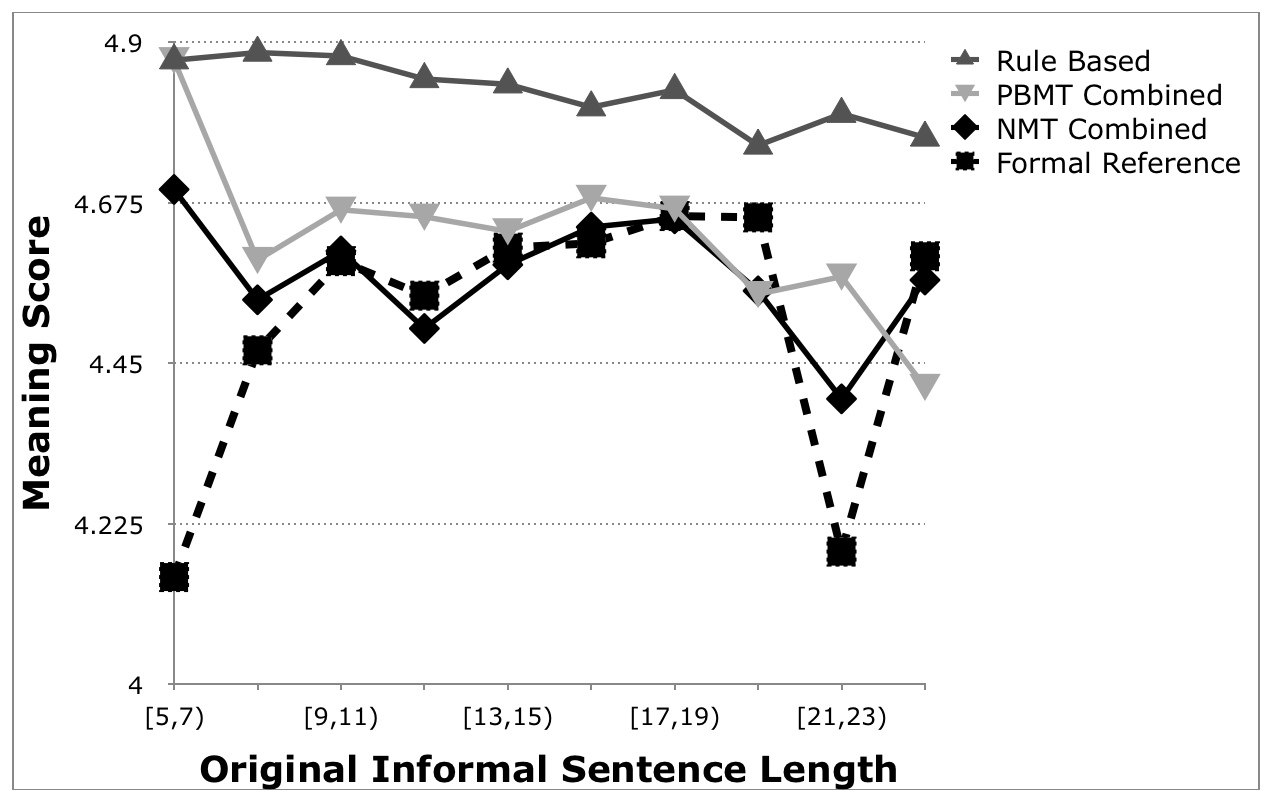}
    \end{minipage}
    \caption{For varying sentence lengths of the original informal sentence the \textit{formality} and  the \textit{meaning} scores from human judgments on different model outputs and on the original informal and the formal reference sentences. }\label{fig:EM_System_Analysis}
\end{figure}



\section{Results}\label{results}

In this section, we discuss how well the five models perform in the \textit{informal} to \textit{formal} style transfer task using human judgments (\S\ref{results_human}) and automatic metrics (\S\ref{results_auto}), the correlation of the automatic metrics and human judgments to determine the efficacy of the metrics (\S\ref{sec:metric_correlation}) and present a manual analysis (\S\ref{manual_analysis}). We randomly select 500 sentences from each test set and run all five models.  We use the entire train and tune split for training and tuning. We discuss results only on the E\&M domain and list results on the F\&R domain in the supplementary material. 

Table ~\ref{tab:results_EM} shows the results for human \S\ref{results_human} and automatic \S\ref{results_auto} evaluation of model rewrites.  For all metrics except \emph{TERp}, a higher score is better. For each of the automatic metrics, we evaluate against four human references. The row \textit{`Original Informal'} contains the scores when the original informal sentence is compared with the four formal reference rewrites. Comparing the model scores to this score helps us understand how closer are the model outputs to the formal reference rewrites compared to initial distance between the informal and the formal reference rewrite.

\subsection{Results using Human Judgments}\label{results_human}
The columns marked \textit{`Human'} in Table~\ref{tab:results_EM} show the human judgments for the models on the three separate criteria of \textit{formality}, \textit{fluency} and \textit{meaning} collected using the process described in Section~\ref{human_judgments}.\footnote{Out of the four reference rewrites, we pick one at random to show to Turkers.} The NMT Baseline and  Copy models beat others on the formality axis by a significant margin. Only the NMT Combined model achieves a statistically higher fluency score when compared to the rule-based baseline model. As expected, the rule-based model is the most meaning preserving since it is the most conservative.  Figure~\ref{fig:EM_System_Analysis} shows the trend in the four leading models along \textit{formality} and \textit{meaning} for varying lengths of the source sentence. NMT Combined beats PBMT on \textit{formality} for shorter lengths whereas the trend reverses as the length increases. PBMT generally preserves meaning more than the NMT Combined. We find that the fluency scores for all models decreases as the sentence length increases which is similar to the trend generally observed with machine translation based approaches. 

Since a good style transfer model is the one that attains a balanced score across all the three axes, we evaluate the models on a combination of these metrics\footnote{We recalibrate the scores to normalize for different ranges.} shown under the column \textit{`Combined'} in Table~\ref{tab:results_EM}. NMT Combined is the only model having a combined score statistically greater than the rule-based approach.

Finally, Table \ref{tab:overall-ranking} shows the overall rankings of the models from best to worst in both domains. PBMT and NMT Combined models beat the rule-based model although not significantly in the E\&M domain but significantly in the F\&R domain. Interestingly, the rule-based approach attains third place with a score significantly higher than NMT Copy and NMT Baseline models. It is important to note here that while such a rule-based approach is relatively easy to craft for the formality style transfer task, the same may not be true for other styles like politeness or persuasiveness.
\begin{table}[htbp]
\small
\begin{tabular}{l | l}
\hline
E\&M &  F\&R \\
\hline\hline
(2.03*) Reference & (2.13*) Reference \\
(2.47) PBMT & (2.38*) PBMT \\
(2.48) NMT Combined & (2.38*) NMT Combined \\
(2.54) Rule-based & (2.56) Rule-based\\
(3.03*) NMT Copy & (2.72*) NMT Copy  \\
(3.03*) NMT Baseline & (2.79*) NMT Baseline \\
\end{tabular}
\caption{Ranking of different models on the \textit{informal} to \textit{formal} style transfer task. Rankings marked with * are significantly different from the rule-based ranking with $p < 0.001$.}\label{tab:overall-ranking}
\vspace*{-3mm}
\end{table}

\begin{table}[htbp]
\centering
\small 
\begin{tabular}{l | c | c | c}
\textbf{Automatic} & \textbf{Human} & \textbf{E\&M} & \textbf{F\&R}\\ 
\hline
Formality & \textit{Formality} & 0.47 & 0.45\\
Fluency & \textit{Fluency} & 0.48 & 0.46\\
Meaning & \textit{Meaning} & 0.33 & 0.30\\
\hline
BLEU & \textit{Overall} & -0.48 & -0.43\\
TERp & \textit{Overall} & 0.31 & 0.30\\
PINC & \textit{Overall} & 0.11 & 0.08\\
\end{tabular}
\caption{Spearman rank correlation between automatic metrics and human judgments. The first three metrics are correlated with their respective human judgments and the last three metrics are correlated with the \textit{overall ranking} human judgments. All correlations are statistically significant with $p < 0.001$.}\label{tab:metric_correlation}
\vspace*{-3mm}
\end{table}

\begin{table*}
\small
\begin{tabular}{l l}
\hline
\hline
\multicolumn{2}{l}{\textbf{Entertainment \& Music}}\\
\hline
\hline
Original Informal & Wow , I am very dumb in my observation skills ......\\
Reference Formal &  I do not have good observation skills . \\
\hline
Rule-based &  Wow , I am very dumb in my observation skills .\\
PBMT &  Wow , I am very dumb in my observation skills .\\
NMT Baseline & I am very foolish in my observation skills .    \\
NMT Copy & Wow , I am very foolish in my observation skills . \\
NMT Combined & I am very unintelligent in my observation skills .\\
\hline
\hline
\multicolumn{2}{l}{\textbf{Family \& Relationship}}\\
\hline
\hline
Original Informal & i hardly everrr see him in school either usually i see hima t my brothers basketball games .\\
Reference Formal & I hardly ever see him in school . I usually see him with my brothers playing basketball . \\
\hline
Rule-based & I hardly everrr see him in school either usually I see hima t my brothers basketball games . \\
PBMT &  I hardly see him in school as well, but my brothers basketball games . \\
NMT  & I rarely see him in school , either I see him at my brother 's basketball games . \\
NMT Copy & I hardly see him in school either , usually I see him at my brother 's basketball games . \\
NMT Combined & I rarely see him in school either usually I see him at my brothers basketball games . \\
\end{tabular}
\caption{Sample model outputs with references from both E\&M and F\&R domains on the \textit{informal} to \textit{formal} task}\label{tab:model_outputs}
\end{table*}

\subsection{Results with Automatic Metrics}\label{results_auto} 
Under automatic metrics, the formality and meaning scores align with the human judgments with the NMT Baseline and NMT Copy winning on formality and rule-based winning on meaning. The fluency score of the NMT Baseline is the highest in contrast to human judgments where the NMT Combined wins. This discrepancy could be due to \acroh being trained on \textit{essays} which contains sentences of a more formal genre compared to Yahoo Answers. In fact, the fluency classifier scores the formal reference quite low as well. Under overall metrics, PBMT  and NMT Combined models beat other models as per BLEU (significantly) and TERp (not significantly). NMT Baseline and NMT copy win over other models as per PINC which can be explained by the fact that PINC measures lexical dissimilarity with the source and NMT models tend towards making more changes. Although such an analysis is useful, for a more thorough understanding of these metrics, we next look at their correlation with human judgments.

\subsection{Metric Correlation}\label{sec:metric_correlation}
We report the spearman rank correlation co-efficient between automatic metrics and human judgments in Table~\ref{tab:metric_correlation}. For \textit{formality}, \textit{fluency} and \textit{meaning}, the correlation is with their respective human judgments whereas for BLEU, TERp and PINC, the correlation is with the overall ranking. 

We see that the formality and the fluency metrics correlate moderately well while the meaning metric correlates comparatively poorly. To be fair, the \acrohe classifier was trained on the STS dataset which contains more formal writing than informal.  BLEU correlates moderately well (better than what \acroxu observed for the Shakespeare task) whereas the correlation drops for TERp. PINC, on the other hand, correlates very poorly with a positive correlation with rank when it should have a negative correlation with rank, just like BLEU. This sheds light on the fact that PINC, on its own, is not a good metric for style transfer since it prefers lexical edits at the cost of meaning changes. In the Shakespeare task, \acroxu did observe a higher correlation with PINC (0.41) although the correlation was not with overall system ranking but rather only on the style metric. Moreover, in the Shakespeare task, changing the text is more favorable than in formality. 

\subsection{Manual Analysis}\label{manual_analysis}
The prior evaluations reveal the relative performance differences between approaches.  Here, we identify trends per and between approaches. We sample 50 informal sentences total from both domains and then analyze the outputs from each model. We present sample sentences in  Table~\ref{tab:model_outputs}.

The NMT Baseline and NMT Copy tend to have the most variance in their performance. This is likely due to the fact that they are trained on only 50K sentence pairs, whereas the other models are trained on much more data. For shorter sentences, these models make some nice formal transformations like from `\textit{very dumb}' to `\textit{very foolish}'.  However, for longer sentences, these models make drastic meaning changes and drop some content altogether (see examples in Table~\ref{tab:model_outputs}). 
On the other hand, the PBMT and NMT Combined models have lower variance in their performance. They make changes more conservatively but when they do, they are usually correct. Thus, most of the outputs from these two models are usually meaning preserving but at the expense of a lower formality score improvement.

In most examples, all models are good at removing very informal words like `\textit{stupid}', `\textit{idiot}' and `\textit{hell}', with PBMT and NMT Combined models doing slightly better.  All models struggle when the original sentence is very informal or disfluent. They all also struggle with sentence completions that humans seem to be very good at. This might be because humans assume a context when absent, whereas the models do not. Unknown tokens, either real words or misspelled words, tend to wreak havoc on all approaches. In most cases, the models simply did not transform that section of the sentence, or remove the unknown tokens. Most models are effective at low-level changes such as writing out numbers, inserting commas, and removing common informal phrases. 

\section{Conclusions and Future Work}
The goal of this paper was to move the field of style transfer forward by 
creating a large training and evaluation corpus to be made public, showing 
that adapting MT techniques to this task can serve as strong baselines for future work, and analyzing the usefulness of existing metrics for overall style transfer as well as three specific criteria of automatic style transfer evaluation.  We view this work as rigorously expanding on the foundation set by \acroxu five years earlier.  It is our hope that with a common test set, the field can finally benchmark approaches which do not require parallel data.

We found that while the NMT systems perform well given automatic metrics, humans had a slight preference for the PBMT approach.  That being said, two of the neural approaches (NMT Baseline and Copy) often made successful changes and larger rewrites that the other
models could not.  However, this often came at the expense of a meaning change.
  
We also introduced new metrics and vetted all metrics using comparison with human judgments.  We found that previously-used metrics did not correlate well with human judgments, and thus should be avoided in system development or final evaluation.  The formality and fluency metrics correlated best and we believe that some combination of these metrics with others would be the best next step in the development of style transfer metrics.  Such a metric could then in turn be used to optimize MT models. Finally, in this work we focused on one particular style, formality. The long term goal is to generalize the methods and metrics to any style.






\section*{Acknowledgments}

The authors would like to thank Yahoo Research for making their data available.
The authors would also like to thank Junchao Zheng and Claudia Leacock for their help in the data creation process, Courtney Napoles for providing the fluency scores, Marcin Junczys-Dowmunt, Rico Sennrich, Ellie Pavlick, Maksym Bezva, Dimitrios Alikaniotis and Kyunghyun Cho for helpful discussion and the three anonymous reviewers for their useful comments and suggestions. 

\newpage

\section{Supplementary material}

In this supplementary material, we add additional details supporting the dataset \S\ref{dataset}, models \S\ref{models} 
and results \S\ref{results} we introduce in the main paper. In \S\ref{results}, we also discuss results of our models on the \textit{formal} to \textit{informal} task.

\subsection{Dataset}\label{dataset}

In Section 3.2 of the main paper, we perform a qualitative analysis to understand the types of edits people made for making a sentence more formal. Table~\ref{tab:rewrite_analysis} shows the frequency of each types of edits and example sentence pairs for the same. In addition, for a subset of the categories for which we can count the edits automatically, we show the frequency of edits on the entire train split of our \acrocorp dataset where we observe a higher percentage of capitalization and punctuation edits as compared to manual counting and a much higher percentage of normalizations. 


\begin{table*}
\footnotesize
\begin{tabular}{l c c l l}
\hline
\textbf{Category} & \textbf{Manual} & \textbf{Auto} & \textbf{Original Informal} & \textbf{Formal Rewrite}\\
\hline
Paraphrase & 47\% & -- &  he iss \textbf{wayyyy hottt} & He is \textbf{very attractive}.\\
Capitalization & 46\% & 51\% &  \textbf{yes}, exept for episode \textbf{iv}. & \textbf{Yes}, but not for episode \textbf{IV}.\\
Punctuation & 40\% & 69\% &  I've watched it and it is AWESOME\textbf{!!!!}& I viewed it and I believe it is a quality \\
& & &  & program\textbf{.}\\
Delete fillers & 26\% & -- & \textbf{Well... }Do you talk to that someone much? & do you talk to that person often?\\
Completion & 15\% & -- & \textbf{Haven't seen} the tv series, but R.O.D. & \textbf{I have not seen} the television series, \\
& & & & however I have seen the R.O.D\\
Spelling & 14\% & -- & that page did not give me \textbf{viroses} (i think) & I don't think that page gave me \textbf{viruses}. \\
Contractions & 12\% & 8\% & I \textbf{didn't} know they had an HBO in the 80's & I \textbf{did not} know HBO existed in the 1980s.\\
Normalization & 10\% & 61\% & my exams \textbf{r} not over yet & My exams \textbf{are} not over yet.\\
Lowercase & 7\% & 8\% & But you will \textbf{DEFINALTELY} know & You will \textbf{definitely} know\\
& & & when you are in love! & when you are in love.\\
Split Sentences & 4\% & -- & it wouldnt be a word, & It would not be a word. \\
& & & it would be me singing operah. & It would be a singing opera. \\
Repetitions & 2\% & 5\% & i'd find out what \textbf{realllllllllllllly} & I would determine what \textbf{really}  \\
& & & happened to marilyn monroe & happened to Marilyn Monroe. \\
\hline
\end{tabular}
\caption{Categories of frequent edits calculated using manual and automatic counting and examples of each. Note that the categories are not mutually exclusive.}\label{tab:rewrite_analysis}
\end{table*}


\begin{table}\label{classifier_correlation}
\footnotesize
\center
\begin{tabular}{c | c | c}
Train & Test & $ Spearman(\rho$) \\ \hline
\acropt & \acropt & 0.68\\ 
\acropt & E\&M & 0.39\\
\acropt & F\&R & 0.38\\
\hline
E\&M & E\&M & 0.56\\
F\&R & F\&R &  0.51\\
\end{tabular}
\caption{Spearman rank correlation between formality classifier predictions and human judgments on 10-fold cross validation on 5000 sentences.}
\vspace*{-3mm}
\end{table}

\subsection{Models}\label{models}
\subsubsection{Rule-based Model}
We use the analysis described in Section 3.2 of the main paper to construct the following set of rules to automatically make an informal sentence more formal: \\
\textbf{Capitalization:} We capitalize the first letter of a sentence, we capitalize the pronoun `I' and we capitalize proper nouns by identifying words with parts of speech NNP or NNPS.\\
\textbf{Lowercase words with all upper cases}: In several informal sentences, words are often capitalized for emphasis, e.g. \textit{``ARE YOU KIDDING ME????''} We lowercase such sentences or words.\\
\textbf{Expand contractions:} Informal sentences contain contractions like \textit{`wasn't', `haven't',} etc. We handcraft a list of expansions for all such contractions.\\
\textbf{Replace slang words:} Informal sentences contain slang word usage like \textit{`juz', `wanna',} etc. We handcraft a list of slang replacements. \\
\textbf{Replace swear words:} Informal sentences frequently contain swear words. We handcraft a list of swear words and replace all but their first character with asterisks. Example, \textit{`suck'} is replaced with \textit{`s***'}.\\
\textbf{Remove character repetition:} Informal sentences contain several instances of repeated characters for emphasis. For example, \textit{`nooooo', `yayyyyy', `!!!!', `???'}. We use regular expressions to replace such repeated occurrences with a single occurrence.

The rule-based model for the second direction of style transfer i.e. from \textit{formal} to \textit{informal} consists of the same rules as above but in the reverse direction. The rules of capitalization, contractions and slang usage are applied always whereas the rules of uppercasing and character repetition are applied proportionally to their occurrences in the \acrocorp train split. 
%
%
\subsubsection{PBMT Model}\label{smt_models}
The different ideas combined together to obtain the PBMT model in Section 4.2 of the main paper are described in detail below: \\\\
\textbf{PBMT on rule-based:} When we use the $\sim$50K sentence pairs in our \acrocorp train set to train a baseline PBMT model, we observe that the model mainly learns to replicate the rules we crafted in our rule-based approach. Hence, in order to force the PBMT model to learn generalizations beyond the rules, we train the PBMT model on the output of our rule-based approach such that the source side of the parallel data is now the output of the rule-based approach. 
For all of our subsequent models we use this parallel data. \\\\
\textbf{Self-training:} The amount of parallel data in our \textit{formality} dataset is orders of magnitude smaller than the amount typically used for training translation models. We therefore increase the size of our train set by way of \textit{self-training} where we use the PBMT model to translate the large number of in-domain sentences from \acrocorp belonging to the the source style and use the resultant output to retrain the PBMT model. \\\\
\textbf{Sub-selection using Edit Distance:} A large portion of the training data obtained via \textit{self-training} consists of parallel sentences where the two sides are almost identical. In order to push the PBMT model towards translations that involve higher number of edits, we sub-select the additional training data generated using \textit{self-training} to include only those where the edit distance between the two sides is more than 10. Further, to ensure the equal proportion of the original parallel data and the additional data, we up-weight the original parallel data via duplication.\\\\ \\
\textbf{Larger Language Model with Data Selection:} 
The Yahoo Answers corpus contains a large number of target style sentences spanning across different domains that we could potentially use to train a larger language model, but at the cost of domain mismatch. To sub sample from large out-of-domain data, we use 
intelligent data selection method and train a language model on sentences that are closer to the target style in-domain data.

Table~\ref{tab:training_data_statistics} contains the approximate sizes of the training data used in the main models across the two domains. Under the ``Combined'' models, the first part is duplication of the \acrocorp train split, the second part is additional data obtained via self-training with sub-selection and the third part is the additional data obtained via back-translation for the ``NMT Combined model''.

\subsubsection{Details of NMT model}
We use the OpenNMT-py \cite{P17-4012} toolkit with default parameters with a vocabulary size of 50K and embeddings of size 300. At test time, we replace unknown tokens with the source token that has the highest attention weight. The input word embeddings are pretrained on Yahoo Answers using GloVE \cite{pennington2014glove}.

\begin{table}[t]
\footnotesize
\begin{tabular}{l l}
Model &  Training data\\
\hline
\hline
& E\&M \\
\hline
PBMT Baseline & 50K\\
PBMT Combined & 50K*6+0.3M\\
NMT Combined & 50K*6+0.3M+0.7M \\
\hline
& F\&R \\
\hline
PBMT Baseline & 50K\\
PBMT Combined & 50K*10+0.3M\\
NMT Combined & 50K*10+0.6M+1.8M\\
\end{tabular}
\caption{Sizes of the training data used in the different models for the two domains.}\label{tab:training_data_statistics}
\end{table}

\begin{table}[htbp]
\small
\begin{tabular}{l | l}
\hline
E\&M &  F\&R \\
\hline\hline
(2.33) PBMT Combined & (2.24) Rule-based \\
(2.40) NMT Combined & (2.28) PBMT Combined \\
(2.44) Rule-based & (2.36) NMT Combined\\
(2.52) Reference & (2.44) Reference \\
(2.69*) NMT Copy & (2.50*) NMT Baseline \\
(2.72*) NMT Baseline & (2.53*) NMT Copy\\
\end{tabular}
\caption{Ranking of different models on \textit{formal} to \textit{informal} task. Rankings marked with * are significantly different from the rule-based ranking with $p < 0.001$.}\label{tab:overall-ranking2}
\end{table}

\label{sec:metric_correlation}
\begin{table}[htbp]
\centering
\small 
\begin{tabular}{l | c | c | c}
\textbf{Automatic} & \textbf{Human} & \textbf{E\&M} & \textbf{F\&R}\\ 
\hline
BLEU & \textit{Overall} & -0.10* & -0.01\\
TERp & \textit{Overall} & 0.06 & 0.02\\
PINC & \textit{Overall} & 0.17* & 0.15* \\
\hline 
Formality & \textit{Formality} & 0.55* & 0.57*\\
Fluency & \textit{Fluency} & 0.46* & 0.43*\\
Meaning & \textit{Meaning} & 0.28* & 0.28*\\
Combined & \textit{Combined} & 0.38* & 0.38*\\
\end{tabular}
\caption{Spearman rank correlation between automatic metrics and human judgments. The first three metrics are correlated with the \textit{overall ranking} human judgments and the last four are correlated with the human judgments on the respective three axes. Correlations marked with * are statistically significant with $p < 0.001$}\label{metric_correlation}
\end{table}

\subsection{Results}\label{results}

\begin{table*}[htbp]
\center
\small
\begin{tabular}{l |c c| c c| c c| c c| c c c}
& \multicolumn{2}{c|}{Formality} & \multicolumn{2}{c|}{Fluency} & \multicolumn{2}{c}{Meaning} & \multicolumn{2}{|c} {Combined} & \multicolumn{3}{|c}{Overall}\\
Model &  Human & PT16 & Human & \acroh & Human & \acrohe & Human & Auto & BLEU & TERp & PINC\\
\hline
\hline
\textit{Original Informal} & \textit{-0.90} & \textit{-0.80} &\textit{3.92} & \textit{3.09} & -- & --  & -- & -- & 52.61 & 0.34 & 0.00\\
Formal Reference & 0.41 & 0.22 & 4.43 & 3.74 & 4.56 & 3.54 & 5.68 & 4.76 & 100.0 & 0.37 & 67.83\\
\hline
Rule-based & -0.18 & -0.15 & 4.08 & 3.26 & \textbf{4.82} & \textbf{4.29} & 5.41 & 4.78 & 68.17 & 0.27 & 26.89\\
PBMT Combined & 0.05* & 0.11* & 4.15 & 3.48* & 4.65* & 4.02* & 5.45 & 4.88* & \textbf{74.32}* & \textbf{0.25}* & 44.77*\\
NMT Baseline & 0.19* & 0.13* & 4.18 & \textbf{3.56}* & 3.88* & 3.91* & 5.20* & \textbf{4.89}* & 69.09* & 0.31* & \textbf{51.00}*\\
NMT Copy & \textbf{0.33}* & \textbf{0.15}* & 4.21* & 3.55* & 3.97* & 3.88* & 5.30 & 4.88* & 69.41 & 0.30 & 50.93\\
NMT Combined & 0.20* & 0.10* & \textbf{4.27}* & 3.45* & 4.69* & 4.06* & \textbf{5.57}* & 4.88* & \textbf{74.60}* & \textbf{0.24}* & 41.52*\\
\hline
\end{tabular}
\caption{Results of models on 500 test sentences for \textit{informal} to \textit{formal} task evaluated using human judgments and automatic metrics for three criteria of evaluation: formality, fluency and meaning preservation on the F\&R domain. Scores marked with * are significantly different from the rule-based scores with $p < 0.001$.}\label{tab:results_FR}
\end{table*}

\subsubsection{Results on F\&R domain}
In the main paper, we include results for only the E\&M domain. Here we discuss the results on the F\&R domain. Table~\ref{tab:results_FR}, similar to Table 4 in the main paper, shows the results of the models on 500 test sentences evaluated using both human judgments and automatic metrics. The main observations regarding model performances across all metrics are similar to the E\&M domain. One difference is that the formality score of the original informal sentences in F\&R are higher than in E\&M and consequently the formality scores of the formal rewrites from both human references and model outputs are higher than in E\&M. 

\begin{table*}[htbp]
\center
\small
\begin{tabular}{l |c c| c c| c c| c c | c c c}
System & \multicolumn{2}{c|}{Formality} & \multicolumn{2}{c|}{Fluency} & \multicolumn{2}{c|}{Meaning} & \multicolumn{2}{c|}{Combined} & \multicolumn{3}{c}{Overall} \\
\hline
&  Human & PT16 & Human & \acroh & Human & \acrohe & Human & Auto & BLEU & TERp & PINC\\
\hline
\hline
\multicolumn{7}{c}{\textbf{Entertainment \& Music}}\\ 
\hline
\hline
\textit{Original Formal} & \textit{0.73} & \textit{0.26} & \textit{4.50} & \textit{3.39} & -- & -- & -- & -- & \textit{29.54} & \textit{0.83} & \textit{0.00}\\
Informal Ref & -0.88 & -0.89 & 4.00 & 3.00 & 3.92 & 3.09 & 0.97 & 0.80 & 100.0 & 0.64 & 81.14\\
\hline
Rule-based & \textbf{-0.96} & \textbf{-0.88} & \textbf{3.92 } & \textbf{2.99} & \textbf{4.80} & 3.95 & \textbf{0.55} & \textbf{0.46} & 22.41 & 0.85 & 47.00\\
PBMT Combined & -0.16* & -0.15* & 4.15* & 3.32* & 4.56* & \textbf{4.28}* & 1.03* & 0.73* & 32.66* & 0.80* & 37.38*\\
NMT Baseline & -0.59* & -0.61* & 3.94 & 3.32* & 3.55* & 3.90 & 1.19* & 0.73* & 29.87* & 0.83* & \textbf{58.38}*\\
NMT Copy & -0.53* & -0.53* & 3.98 & 3.30* & 3.53* & 3.96 & 1.24* & 0.72* & 31.29* & 0.80* & 55.42*\\
NMT Combined & -0.34* & -0.39* & 4.08* & 3.27* & 4.32* & 4.12* & 1.03* & 0.69* & \textbf{34.07}* & \textbf{0.78}* & 42.31\\
\hline
\hline
\multicolumn{7}{c}{\textbf{Family \& Relationships}}\\ 
\hline
\hline
\textit{Original Formal} & \textit{0.73} & \textit{0.36} & \textit{4.47} & \textit{3.49} & -- & -- & -- & -- & \textit{28.64} & \textit{0.84} & \textit{0.00}\\
Informal Ref & -0.60 & -0.87 & 4.09 & 3.21 & 3.82 & 3.25 & 1.15 & 0.85 & 100.0 & 0.65 & 80.69\\
\hline
Rule-based & \textbf{-1.05} & \textbf{-0.94} & \textbf{3.89} & \textbf{2.93} & \textbf{4.46} & 3.98 & \textbf{0.65} & \textbf{0.40} & 19.56 & 0.87 & 45.95\\
PBMT Combined & -0.17* & -0.19* & 4.18* & 3.36* & 4.45 & \textbf{4.43}* & 1.09* & 0.68* & 31.20* & 0.82* & 26.74*\\
NMT Baseline & -0.16* & -0.46* & 4.08* & 3.34* & 3.80* & 4.11 & 1.30* & 0.71* & \textbf{34.69}* & \textbf{0.78}* & \textbf{48.68}\\
NMT Copy & -0.19* & -0.36* & 4.08* & 3.34* & 3.74* & 4.14 & 1.32* & 0.73* & 33.76* & \textbf{0.78}* & 47.52\\
NMT Combined & -0.12* & -0.34* & 4.17* & 3.33* & 4.22* & 4.26* & 1.19* & 0.69* & 33.57* & 0.79* & 36.76\\
\hline
\end{tabular}
\caption{Results of models on 500 test sentences for \textit{formal} to \textit{informal} task evaluated using human judgments and automatic metrics  for three criteria of evaluation: formality, fluency and meaning preservation on E\&M and F\&R domains. Scores marked with * are significantly different from the rule-based scores with $p < 0.001$.}\label{tab:results_formal_to_informal}
\end{table*}

\subsubsection{Results on \textit{Formal} to \textit{Informal} task}
In the main paper, we focus on the \textit{informal} to \textit{formal} direction of style transfer. In this section we discuss the results of our models on the other direction. 
It should be noted that this experimentation is fundamentally different from the first direction in a way that instead of identifying \textit{formal} sentences from Yahoo Answers and collecting their \textit{informal} rewrites, we reuse the data created for the first direction. 

In Table~\ref{tab:results_formal_to_informal}, we show the results of the five main models on the \textit{formal} to \textit{informal} task. The main observation is that, in contrast to the first direction, the rule-based model beats all other models across all three criteria of \textit{formality}, \textit{fluency} and \textit{meaning} as per human judgments and automatic metrics (with the exception of meaning automatic metric where PBMT Combined beats rule-based). NMT Combined and PBMT Combined win as per BLEU and TERp and NMT Baseline wins as per PINC. As in Section 6.3 of the main paper, in Table~\ref{metric_correlation}, we report the correlation of these metrics with human judgments. In contrast to the first direction, the formality classifier obtains a higher correlation which might be because the classifier is trained on informal data and so it is better at assessing informal model outputs than formal model outputs. The fluency and meaning correlation are about the same. In contrast, BLEU, TERp and PINC all three correlate very poorly with the overall ranking. This difference  might be explained by the fact that informal reference rewrites vary highly in that there are much higher number of ways of making a sentence more informal as compared to making it more formal. Therefore, metrics that make use of references might be ill-suited for this style transfer task.

In Table~\ref{tab:model_outputs}, we show some sample model outputs for the E\&M and F\&M domain sentences. We can see that rule-based method uses simple lexical transformations like `just' to `juz', `you' to `u', `because' to `cuz', `love' to `luv', etc and wins over other models. The intent of evaluating our models on this direction of task was to understand how well the same model would do the reverse task. We find that the second direction has different set of challenges and requires models that cater to those specifically if we wish to beat simple rule-based methods.

\begin{table*}
\small
\begin{tabular}{l l}
\hline
\hline
\multicolumn{2}{l}{\textbf{Entertainment \& Music}}\\
\hline
\hline
Original Formal & I am just glad they didn 't show us the toilets .\\
Reference Informal & IM GLAD THEY PASSED THE TOILETS \\
Rule-based & i am juz glad they didn 't show us the toilets .....\\
PBMT Combined & I am just glad they didnt show us the toilets .\\
NMT Baseline & I 'm just glad they didn 't show us the restroom .\\
NMT Copy & I 'm just glad they didn 't show us the brids . \\
NMT Combined & I 'm just glad they didn 't show us the toilets . \\
\hline
\hline
\multicolumn{2}{l}{\textbf{Family \& Relationship}}\\
\hline
\hline
Original Formal & Hopefully , you married your husband because you love him .\\
Reference Informal & you married your hubby hopefully because you love him . \\
Rule-based & hopefully , u MARRIED ur husband coz u luv him .....\\
PBMT Combined & hopefully , you married your husband because you love him .\\
NMT Baseline & you married your husband because you love him .\\
NMT Copy & Hopefully you married your husband because you love him .\\
NMT Combined & Hopefully you married your husband because you love him .\\
\end{tabular}
\caption{Sample model outputs with references from both E\&M and F\&R domains on the \textit{formal} to \textit{informal} task}\label{tab:model_outputs}
\end{table*}

\bibliography{style_transfer}

\begin{thebibliography}{}
\expandafter\ifx\csname natexlab\endcsname\relax\def\natexlab#1{#1}\fi

\bibitem[{Agirre et~al.(2016)Agirre, Banea, Cer, Diab, Gonzalez-Agirre,
  Mihalcea, Rigau, and Wiebe}]{agirre2016semeval}
Eneko Agirre, Carmen Banea, Daniel~M Cer, Mona~T Diab, Aitor Gonzalez-Agirre,
  Rada Mihalcea, German Rigau, and Janyce Wiebe. 2016.
\newblock Semeval-2016 task 1: Semantic textual similarity, monolingual and
  cross-lingual evaluation.
\newblock In {\em SemEval@ NAACL-HLT\/}. pages 497--511.

\bibitem[{Bahdanau et~al.(2014)Bahdanau, Cho, and Bengio}]{bahdanau2014neural}
Dzmitry Bahdanau, Kyunghyun Cho, and Yoshua Bengio. 2014.
\newblock Neural machine translation by jointly learning to align and
  translate.
\newblock {\em arXiv preprint arXiv:1409.0473\/} .

\bibitem[{Brooke and Hirst(2014)}]{brooke2014supervised}
Julian Brooke and Graeme Hirst. 2014.
\newblock Supervised ranking of co-occurrence profiles for acquisition of
  continuous lexical attributes.
\newblock In {\em COLING\/}. pages 2172--2183.

\bibitem[{Brooke et~al.(2010)Brooke, Wang, and Hirst}]{brooke2010automatic}
Julian Brooke, Tong Wang, and Graeme Hirst. 2010.
\newblock Automatic acquisition of lexical formality.
\newblock In {\em Proceedings of the 23rd International Conference on
  Computational Linguistics: Posters\/}. Association for Computational
  Linguistics, pages 90--98.

\bibitem[{Callison-Burch(2008)}]{callisonburch:2008:EMNLP}
Chris Callison-Burch. 2008.
\newblock \href{http://www.aclweb.org/anthology/D08-1021}{Syntactic constraints
  on paraphrases extracted from parallel corpora}.
\newblock In {\em Proceedings of the 2008 Conference on Empirical Methods in
  Natural Language Processing\/}. Association for Computational Linguistics,
  Honolulu, Hawaii, pages 196--205.
\newblock \url{http://www.aclweb.org/anthology/D08-1021}.

\bibitem[{Chen and Dolan(2011)}]{chen2011collecting}
David~L Chen and William~B Dolan. 2011.
\newblock Collecting highly parallel data for paraphrase evaluation.
\newblock In {\em Proceedings of the 49th Annual Meeting of the Association for
  Computational Linguistics: Human Language Technologies-Volume 1\/}.
  Association for Computational Linguistics, pages 190--200.

\bibitem[{Cho et~al.(2014)Cho, van Merri{\"e}nboer, Bahdanau, and
  Bengio}]{cho2014properties}
Kyunghyun Cho, Bart van Merri{\"e}nboer, Dzmitry Bahdanau, and Yoshua Bengio.
  2014.
\newblock On the properties of neural machine translation: Encoder--decoder
  approaches.
\newblock {\em Syntax, Semantics and Structure in Statistical Translation\/}
  page 103.

\bibitem[{Ficler and Goldberg(2017)}]{ficler2017controlling}
Jessica Ficler and Yoav Goldberg. 2017.
\newblock Controlling linguistic style aspects in neural language generation.
\newblock {\em Proceedings of the Workshop on Stylistic Variation, EMNLP
  2017\/} .

\bibitem[{He et~al.(2015)He, Gimpel, and Lin}]{he2015multi}
Hua He, Kevin Gimpel, and Jimmy~J Lin. 2015.
\newblock Multi-perspective sentence similarity modeling with convolutional
  neural networks.
\newblock In {\em EMNLP\/}. pages 1576--1586.

\bibitem[{Heafield et~al.(2013)Heafield, Pouzyrevsky, Clark, and
  Koehn}]{Heafield-estimate}
Kenneth Heafield, Ivan Pouzyrevsky, Jonathan~H. Clark, and Philipp Koehn. 2013.
\newblock
  \href{https://kheafield.com/papers/edinburgh/estimate\_paper.pdf}{Scalable
  modified {Kneser-Ney} language model estimation}.
\newblock In {\em Proceedings of the 51st Annual Meeting of the Association for
  Computational Linguistics\/}. Sofia, Bulgaria, pages 690--696.
\newblock \url{https://kheafield.com/papers/edinburgh/estimate\_paper.pdf}.

\bibitem[{Heilman et~al.(2014)Heilman, Cahill, Madnani, Lopez, Mulholland, and
  Tetreault}]{heilman-EtAl:2014:P14-2}
Michael Heilman, Aoife Cahill, Nitin Madnani, Melissa Lopez, Matthew
  Mulholland, and Joel Tetreault. 2014.
\newblock \href{http://www.aclweb.org/anthology/P14-2029}{Predicting
  grammaticality on an ordinal scale}.
\newblock In {\em Proceedings of the 52nd Annual Meeting of the Association for
  Computational Linguistics (Volume 2: Short Papers)\/}. Association for
  Computational Linguistics, Baltimore, Maryland, pages 174--180.
\newblock \url{http://www.aclweb.org/anthology/P14-2029}.

\bibitem[{Heylighen and Dewaele(1999)}]{heylighen1999formality}
Francis Heylighen and Jean-Marc Dewaele. 1999.
\newblock Formality of language: definition, measurement and behavioral
  determinants.
\newblock {\em Interner Bericht, Center “Leo Apostel”, Vrije Universiteit
  Br{\"u}ssel\/} .

\bibitem[{Hochreiter and Schmidhuber(1997)}]{hochreiter1997long}
Sepp Hochreiter and J{\"u}rgen Schmidhuber. 1997.
\newblock Long short-term memory.
\newblock {\em Neural computation\/} 9(8):1735--1780.

\bibitem[{Hovy(1987)}]{hovy1987generating}
Eduard Hovy. 1987.
\newblock Generating natural language under pragmatic constraints.
\newblock {\em Journal of Pragmatics\/} 11(6):689--719.

\bibitem[{Hu et~al.(2017)Hu, Yang, Liang, Salakhutdinov, and
  Xing}]{hu2017toward}
Zhiting Hu, Zichao Yang, Xiaodan Liang, Ruslan Salakhutdinov, and Eric~P Xing.
  2017.
\newblock Toward controlled generation of text.
\newblock In {\em International Conference on Machine Learning\/}. pages
  1587--1596.

\bibitem[{Jhamtani et~al.(2017)Jhamtani, Gangal, Hovy, and
  Nyberg}]{jhamtani2017shakespearizing}
Harsh Jhamtani, Varun Gangal, Eduard Hovy, and Eric Nyberg. 2017.
\newblock Shakespearizing modern language using copy-enriched
  sequence-to-sequence models.
\newblock {\em Proceedings of the Workshop on Stylistic Variation, EMNLP
  2017\/} pages 10--19.

\bibitem[{Kajiwara and Komachi(2016)}]{kajiwara2016building}
Tomoyuki Kajiwara and Mamoru Komachi. 2016.
\newblock Building a monolingual parallel corpus for text simplification using
  sentence similarity based on alignment between word embeddings.
\newblock In {\em Proceedings of COLING 2016, the 26th International Conference
  on Computational Linguistics: Technical Papers\/}. pages 1147--1158.

\bibitem[{Klein et~al.(2017)Klein, Kim, Deng, Senellart, and Rush}]{P17-4012}
Guillaume Klein, Yoon Kim, Yuntian Deng, Jean Senellart, and Alexander Rush.
  2017.
\newblock \href{https://doi.org/10.18653/v1/P17-4012}{Opennmt: Open-source
  toolkit for neural machine translation}.
\newblock In {\em Proceedings of ACL 2017, System Demonstrations\/}.
  Association for Computational Linguistics, pages 67--72.
\newblock \url{https://doi.org/10.18653/v1/P17-4012}.

\bibitem[{Koehn et~al.(2007)Koehn, Hoang, Birch, Callison-Burch, Federico,
  Bertoldi, Cowan, Shen, Moran, Zens et~al.}]{koehn2007moses}
Philipp Koehn, Hieu Hoang, Alexandra Birch, Chris Callison-Burch, Marcello
  Federico, Nicola Bertoldi, Brooke Cowan, Wade Shen, Christine Moran, Richard
  Zens, et~al. 2007.
\newblock Moses: Open source toolkit for statistical machine translation.
\newblock In {\em Proceedings of the 45th annual meeting of the ACL on
  interactive poster and demonstration sessions\/}. Association for
  Computational Linguistics, pages 177--180.

\bibitem[{Lahiri et~al.(2011)Lahiri, Mitra, and Lu}]{lahiri2011informality}
Shibamouli Lahiri, Prasenjit Mitra, and Xiaofei Lu. 2011.
\newblock Informality judgment at sentence level and experiments with formality
  score.
\newblock In {\em International Conference on Intelligent Text Processing and
  Computational Linguistics\/}. Springer, pages 446--457.

\bibitem[{Moore and Lewis(2010)}]{moore2010intelligent}
Robert~C Moore and William Lewis. 2010.
\newblock Intelligent selection of language model training data.
\newblock In {\em Proceedings of the ACL 2010 conference short papers\/}.
  Association for Computational Linguistics, pages 220--224.

\bibitem[{Mosquera and Moreda(2012)}]{mosquera2012smile}
Alejandro Mosquera and Paloma Moreda. 2012.
\newblock Smile: An informality classification tool for helping to assess
  quality and credibility in web 2.0 texts.
\newblock In {\em Proceedings of the ICWSM workshop: Real-Time Analysis and
  Mining of Social Streams (RAMSS)\/}.

\bibitem[{Napoles et~al.(2016)Napoles, Sakaguchi, and
  Tetreault}]{napoles-sakaguchi-tetreault:2016:EMNLP2016}
Courtney Napoles, Keisuke Sakaguchi, and Joel Tetreault. 2016.
\newblock \href{https://aclweb.org/anthology/D16-1228}{There's no comparison:
  Reference-less evaluation metrics in grammatical error correction}.
\newblock In {\em Proceedings of the 2016 Conference on Empirical Methods in
  Natural Language Processing\/}. Association for Computational Linguistics,
  Austin, Texas, pages 2109--2115.
\newblock \url{https://aclweb.org/anthology/D16-1228}.

\bibitem[{Niu et~al.(2017)Niu, Martindale, and Carpuat}]{niu2017study}
Xing Niu, Marianna Martindale, and Marine Carpuat. 2017.
\newblock A study of style in machine translation: Controlling the formality of
  machine translation output.
\newblock In {\em Proceedings of the 2017 Conference on Empirical Methods in
  Natural Language Processing\/}. pages 2804--2809.

\bibitem[{Papineni et~al.(2002)Papineni, Roukos, Ward, and
  Zhu}]{papineni2002bleu}
Kishore Papineni, Salim Roukos, Todd Ward, and Wei-Jing Zhu. 2002.
\newblock Bleu: a method for automatic evaluation of machine translation.
\newblock In {\em Proceedings of the 40th annual meeting on association for
  computational linguistics\/}. Association for Computational Linguistics,
  pages 311--318.

\bibitem[{Pavlick and Nenkova(2015)}]{pavlick2015inducing}
Ellie Pavlick and Ani Nenkova. 2015.
\newblock Inducing lexical style properties for paraphrase and genre
  differentiation.
\newblock In {\em HLT-NAACL\/}. pages 218--224.

\bibitem[{Pavlick and Tetreault(2016)}]{pavlick2016empirical}
Ellie Pavlick and Joel Tetreault. 2016.
\newblock An empirical analysis of formality in online communication.
\newblock {\em Transactions of the Association for Computational Linguistics\/}
  4:61--74.

\bibitem[{Pennington et~al.(2014)Pennington, Socher, and
  Manning}]{pennington2014glove}
Jeffrey Pennington, Richard Socher, and Christopher Manning. 2014.
\newblock Glove: Global vectors for word representation.
\newblock In {\em Proceedings of the 2014 conference on empirical methods in
  natural language processing (EMNLP)\/}. pages 1532--1543.

\bibitem[{Peterson et~al.(2011)Peterson, Hohensee, and Xia}]{peterson2011email}
Kelly Peterson, Matt Hohensee, and Fei Xia. 2011.
\newblock Email formality in the workplace: A case study on the enron corpus.
\newblock In {\em Proceedings of the Workshop on Languages in Social Media\/}.
  Association for Computational Linguistics, pages 86--95.

\bibitem[{Sennrich et~al.(2016{\natexlab{a}})Sennrich, Haddow, and
  Birch}]{sennrich2016controlling}
Rico Sennrich, Barry Haddow, and Alexandra Birch. 2016{\natexlab{a}}.
\newblock Controlling politeness in neural machine translation via side
  constraints.
\newblock In {\em HLT-NAACL\/}. pages 35--40.

\bibitem[{Sennrich et~al.(2016{\natexlab{b}})Sennrich, Haddow, and
  Birch}]{sennrich-haddow-birch:2016:WMT}
Rico Sennrich, Barry Haddow, and Alexandra Birch. 2016{\natexlab{b}}.
\newblock \href{http://www.aclweb.org/anthology/W16-2323}{Edinburgh neural
  machine translation systems for wmt 16}.
\newblock In {\em Proceedings of the First Conference on Machine
  Translation\/}. Association for Computational Linguistics, Berlin, Germany,
  pages 371--376.
\newblock \url{http://www.aclweb.org/anthology/W16-2323}.

\bibitem[{Sennrich et~al.(2016{\natexlab{c}})Sennrich, Haddow, and
  Birch}]{P16-1009}
Rico Sennrich, Barry Haddow, and Alexandra Birch. 2016{\natexlab{c}}.
\newblock \href{https://doi.org/10.18653/v1/P16-1009}{Improving neural machine
  translation models with monolingual data}.
\newblock In {\em Proceedings of the 54th Annual Meeting of the Association for
  Computational Linguistics (Volume 1: Long Papers)\/}. Association for
  Computational Linguistics, pages 86--96.
\newblock \url{https://doi.org/10.18653/v1/P16-1009}.

\bibitem[{Sheikha and Inkpen(2010)}]{sheikha2010automatic}
Fadi~Abu Sheikha and Diana Inkpen. 2010.
\newblock Automatic classification of documents by formality.
\newblock In {\em Natural Language Processing and Knowledge Engineering
  (NLP-KE), 2010 International Conference on\/}. IEEE, pages 1--5.

\bibitem[{Sheikha and Inkpen(2011)}]{sheikha2011generation}
Fadi~Abu Sheikha and Diana Inkpen. 2011.
\newblock Generation of formal and informal sentences.
\newblock In {\em Proceedings of the 13th European Workshop on Natural Language
  Generation\/}. Association for Computational Linguistics, pages 187--193.

\bibitem[{Snover et~al.(2009)Snover, Madnani, Dorr, and
  Schwartz}]{snover2009fluency}
Matthew Snover, Nitin Madnani, Bonnie~J Dorr, and Richard Schwartz. 2009.
\newblock Fluency, adequacy, or hter?: exploring different human judgments with
  a tunable mt metric.
\newblock In {\em Proceedings of the Fourth Workshop on Statistical Machine
  Translation\/}. Association for Computational Linguistics, pages 259--268.

\bibitem[{Sutskever et~al.(2014)Sutskever, Vinyals, and
  Le}]{sutskever2014sequence}
Ilya Sutskever, Oriol Vinyals, and Quoc~V Le. 2014.
\newblock Sequence to sequence learning with neural networks.
\newblock In {\em Advances in neural information processing systems\/}. pages
  3104--3112.

\bibitem[{Ueffing(2006)}]{ueffing2006self}
Nicola Ueffing. 2006.
\newblock Self-training for machine translation.
\newblock In {\em NIPS workshop on Machine Learning for Multilingual
  Information Access\/}.

\bibitem[{Wang et~al.(2016)Wang, Chen, Rochford, and Qiang}]{wang2016text}
Tong Wang, Ping Chen, John Rochford, and Jipeng Qiang. 2016.
\newblock Text simplification using neural machine translation.
\newblock In {\em AAAI\/}. pages 4270--4271.

\bibitem[{Wubben et~al.(2012{\natexlab{a}})Wubben, Van Den~Bosch, and
  Krahmer}]{wubben2012sentence}
Sander Wubben, Antal Van Den~Bosch, and Emiel Krahmer. 2012{\natexlab{a}}.
\newblock Sentence simplification by monolingual machine translation.
\newblock In {\em Proceedings of the 50th Annual Meeting of the Association for
  Computational Linguistics: Long Papers-Volume 1\/}. Association for
  Computational Linguistics, pages 1015--1024.

\bibitem[{Wubben et~al.(2012{\natexlab{b}})Wubben, van~den Bosch, and
  Krahmer}]{wubben-vandenbosch-krahmer:2012:ACL2012}
Sander Wubben, Antal van~den Bosch, and Emiel Krahmer. 2012{\natexlab{b}}.
\newblock \href{http://www.aclweb.org/anthology/P12-1107}{Sentence
  simplification by monolingual machine translation}.
\newblock In {\em Proceedings of the 50th Annual Meeting of the Association for
  Computational Linguistics (Volume 1: Long Papers)\/}. Association for
  Computational Linguistics, Jeju Island, Korea, pages 1015--1024.
\newblock \url{http://www.aclweb.org/anthology/P12-1107}.

\bibitem[{Xu et~al.(2016)Xu, Napoles, Pavlick, Chen, and
  Callison-Burch}]{xu2016optimizing}
Wei Xu, Courtney Napoles, Ellie Pavlick, Quanze Chen, and Chris Callison-Burch.
  2016.
\newblock Optimizing statistical machine translation for text simplification.
\newblock {\em Transactions of the Association for Computational Linguistics\/}
  4:401--415.

\bibitem[{Xu et~al.(2012)Xu, Ritter, Dolan, Grishman, and
  Cherry}]{xu2012paraphrasing}
Wei Xu, Alan Ritter, Bill Dolan, Ralph Grishman, and Colin Cherry. 2012.
\newblock Paraphrasing for style.
\newblock {\em Proceedings of COLING 2012\/} pages 2899--2914.

\bibitem[{Zhu et~al.(2010)Zhu, Bernhard, and Gurevych}]{zhu2010monolingual}
Zhemin Zhu, Delphine Bernhard, and Iryna Gurevych. 2010.
\newblock A monolingual tree-based translation model for sentence
  simplification.
\newblock In {\em Proceedings of the 23rd international conference on
  computational linguistics\/}. Association for Computational Linguistics,
  pages 1353--1361.

\end{thebibliography}
\bibliographystyle{acl_natbib}

\end{document}